\documentclass[conference]{IEEEtran}
\IEEEoverridecommandlockouts
\usepackage{cite}
\usepackage{amsmath,amssymb,amsfonts}
\usepackage{algorithmic}
\usepackage{graphicx}
\usepackage{textcomp}
\usepackage{xcolor}
\usepackage{tikz}
\usepackage{subcaption}
\usepackage{multirow}
\usepackage[hidelinks]{hyperref}
\usepackage[all]{hypcap}
\usepackage{placeins}
\DeclareMathOperator*{\argmin}{arg\,min}
\usetikzlibrary{matrix,positioning,shapes.geometric,arrows.meta}
\usetikzlibrary{arrows.meta,positioning,shapes.geometric}

\def\BibTeX{{\rm B\kern-.05em{\sc i\kern-.025em b}\kern-.08em
    T\kern-.1667em\lower.7ex\hbox{E}\kern-.125emX}}

\begin{document}

\title{Dual-Arm Tabletop Rearrangement with Movable Stacking and Gap-Aware Push Transfers}

\author{
    \IEEEauthorblockN{
        Mohammadreza Firoozalizadeh,
        Zakiye Rostamirad,
        Arman Barghi,\\
        Ahmad Kalhor,
        Mohammadreza Nayeri,
        Mehdi Tale Masouleh
    }
    \IEEEauthorblockA{\textit{Human and Robot Interaction Laboratory (TaarLab)} \\
    \textit{School of Electrical and Computer Engineering} \\
    \textit{University of Tehran} \\
    Tehran, Iran}
    \IEEEauthorblockA{
        Email: \{m.reza.alizadeh, zakiye.rostamirad, arman.barghi, akalhor, m.nayeri, m.t.masouleh\}@ut.ac.ir
    }
    \thanks{%
        Accepted author manuscript of:
        M. Firoozalizadeh, Z. Rostamirad, A. Barghi,
        A. Kalhor, M. Nayeri, and M. Tale Masouleh,
        ``Dual-Arm Tabletop Rearrangement with Movable
        Stacking and Gap-Aware Push Transfers,''
        in \emph{Proc. 34th International Conference on
        Electrical Engineering (ICEE)}, 2026.
        DOI: \href{https://doi.org/10.1109/ICEE70628.2026.11710769}%
        {10.1109/ICEE70628.2026.11710769}.
        \newline
        \textcopyright{} 2026 IEEE.
        Personal use of this material is permitted.
        Permission from IEEE must be obtained for all other
        uses, in any current or future media, including
        reprinting/republishing this material for advertising
        or promotional purposes, creating new collective
        works, for resale or redistribution to servers or
        lists, or reuse of any copyrighted component of
        this work in other works.
    }%
}

\maketitle

\begin{abstract}
    Object rearrangement in cluttered tabletop scenes is challenging, especially in wide workspaces where dual-arm systems improve coverage and efficiency. Limited free space often necessitates buffering, while objects near or across the shared reachability region of the manipulators must be transferred under careful coordination to avoid inter-arm collisions. Conventional inter-arm transfer typically relies on a fixed handoff configuration, which constrains where and how objects are exchanged, enforces spatiotemporal synchronization of the arms, and reduces flexibility in dense layouts. This paper presents a dual-arm rearrangement framework that combines movable stacking with Gap-Aware Push Transfer for asynchronous cross-boundary exchange. The method scans the shared boundary to identify collision-free transfer gaps, places an object near a selected crossing point, and then pushes it into the receiving workspace without requiring simultaneous interaction between the two arms. A joint Monte Carlo Tree Search (MCTS)-based planner coordinates both manipulators while reasoning over buffering and transfer decisions, and execution cost is used after planning to assess the resulting plan. Experiments show that the proposed method shortens joint execution sequences and becomes increasingly advantageous over a fixed handoff baseline as scene density increases, achieving lower execution cost in denser scenes while remaining competitive in lighter ones.
\end{abstract}


\section{INTRODUCTION}
\begin{figure}[t]
    \centering
    \includegraphics[width=0.98\linewidth]{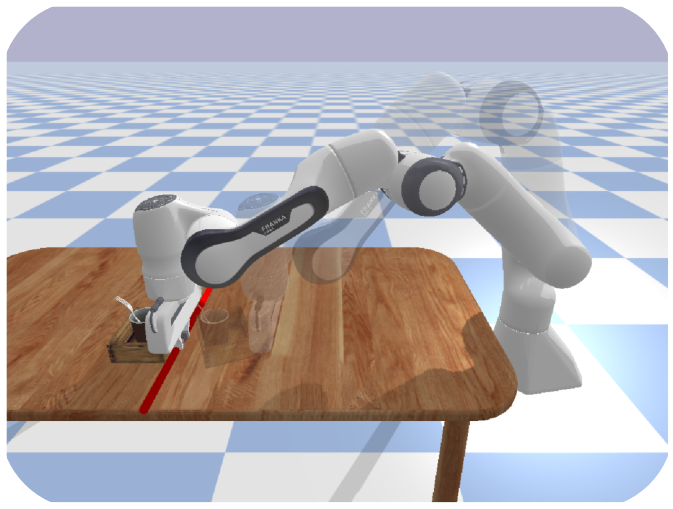}
    \caption{Overview of the proposed dual-arm rearrangement framework, where temporary stacks are formed near the shared boundary for movable stacking~\cite{barghi2025dynamic}, and objects are transferred between workspaces.}
    \label{fig:overview}
\end{figure}

The ability of an autonomous agent to physically interact with its environment and modify object configurations is central to embodied AI~\cite{batra2020rearrangement}. Object rearrangement is a canonical task in this area because it couples perception, planning, and contact-rich manipulation under geometric constraints. In tabletop settings, the problem becomes especially challenging in cluttered scenes, where limited free space and overlap between start and goal poses often require intermediate buffers and multi-step relocations~\cite{han2018complexity, krontiris2016efficiently}.

Object rearrangement in clutter is inherently combinatorial because resolving occlusions among multiple objects requires reasoning over action sequences rather than single-step feasibility~\cite{han2018complexity}. Search-based planners have therefore been widely studied for cluttered manipulation~\cite{krontiris2016efficiently}. Among them, Monte Carlo Tree Search (MCTS) balances exploration and exploitation~\cite{kocsis2006bandit} and has shown strong performance on large discrete rearrangement action spaces~\cite{labbe2020monte, gao2022toward}.

Extending rearrangement planning to dual-arm systems introduces an additional coordination challenge. Although two manipulators can improve efficiency through parallelism, they also create tighter coupling in shared workspaces: an action useful for one arm may obstruct the other, and naive coordination can cause deadlocks when free space is scarce~\cite{abbas2023systematic}. These interactions become more severe in dense layouts, where each relocation affects later action feasibility.

\begin{figure}[t]
\centering
\includegraphics[width=0.98\linewidth]{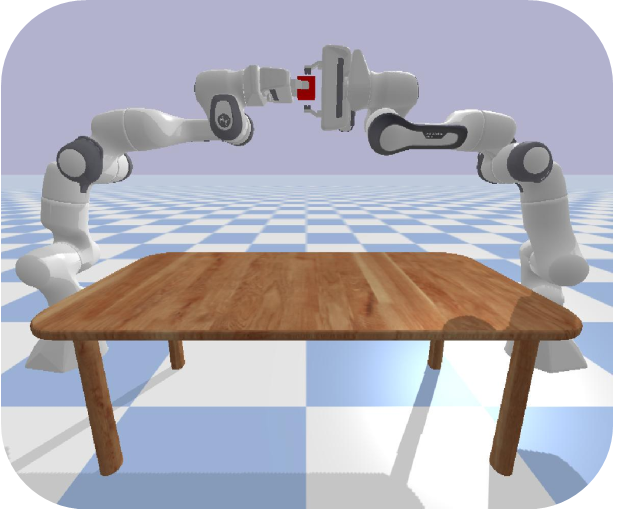}
\caption{Illustration of a fixed aerial handoff configuration for inter-arm transfer in dual-arm rearrangement, adapted from~\cite{gao2022toward}.}
\label{fig:baseline}
\end{figure}

Many dual-arm methods rely on heuristic coordination, dependency reasoning, or task-and-motion planning frameworks~\cite{gao2022toward, shome2021fast, schoels2020decoupled}. In many of them, as illustrated in Fig.~\ref{fig:baseline}, inter-arm exchange is handled through a fixed handoff configuration, restricting where objects can cross the shared boundary and often forcing both manipulators to meet at a specific location within overlapping reach. This coupling can reduce transfer flexibility, especially in cluttered scenes. As summarized in Table~\ref{table:comparison}, representative prior frameworks differ not only in robot setting and planning backbone, but also in coordination style, transfer mechanism, and the assumptions underlying rearrangement execution. To relax this limitation, asynchronous planning and execution have also been explored in related multi-robot manipulation settings~\cite{xu2025apex}. Although such approaches are not developed specifically for tabletop rearrangement, they highlight the potential benefit of reducing strict synchronization between agents. These designs increase flexibility, but in tightly constrained rearrangement transfer, decisions remain strongly coupled with buffering and placement.

Non-prehensile manipulation, particularly pushing, has long been used to extend reachability and enable object motion in cluttered environments~\cite{dogar2011framework, lynch1996stable}. In the present setting, pushing can facilitate cross-boundary object exchange without requiring a synchronous grasp-and-release handoff between the two manipulators. However, in many frameworks the shared boundary is still treated as a predetermined transfer interface rather than as a region whose crossing location can be selected online.

Beyond choosing where and how objects cross the shared boundary, non-monotone rearrangement also requires effective management of intermediate buffers. Most dual-arm planners model buffers as static planar table locations~\cite{shome2021fast}, which is restrictive in densely cluttered scenes. Prior work has shown that stacking can compact objects, free workspace, and serve as temporary buffering during rearrangement~\cite{lee2022high, gao2024orla}. Recent work further shows that temporary stacks can remain movable rather than acting only as strictly static structures~\cite{barghi2025dynamic}. In the present setting, this is valuable because multiple objects can occupy a smaller footprint, improving local space utilization and creating better conditions for inter-arm transfer.

\begin{figure}[t]
\centering
\includegraphics[width=\linewidth]{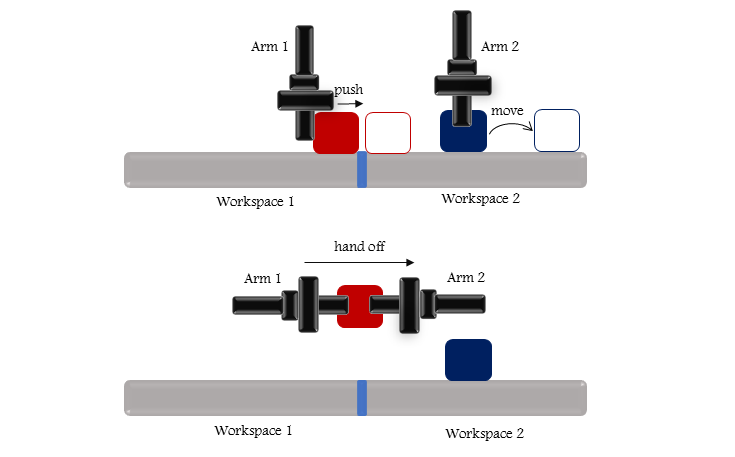}
\caption{Asynchronous cross-boundary push transfer (proposed method) versus synchronous fixed handoff.}
\label{fig:async_transfer}
\end{figure}

In fixed handoff formulations, inter-arm exchange requires both arms to meet at a designated transfer configuration, which tightly couples the timing of the two manipulators. In contrast, a push-based transfer strategy relaxes this synchronization requirement by allowing one arm to push an object across the shared boundary while the other arm continues its local operation and retrieves the object later, as depicted in Fig.~\ref{fig:async_transfer}. The challenge is therefore not only to choose a transfer primitive, but also to integrate transfer selection with buffering and dual-arm coordination under clutter.

\begin{table*}[t]
\caption{\centering Qualitative comparison of representative rearrangement frameworks in terms of robot setup, planning backbone, buffering model, and inter-arm exchange/transfer mechanism.}
\label{table:comparison}
\centering
\resizebox{\textwidth}{!}{%
\begin{tabular}{|l|c|c|c|c|}
\hline
\textbf{Reference} & \textbf{Robot Setup} & \textbf{Planning Backbone} & \textbf{Buffering Model} & \textbf{Inter-arm Exchange / Transfer} \\ \hline

Gao et al.~\cite{gao2022toward}
& Dual-arm
& Task planning
& Lazy buffers
& Predefined handoff pose
\\ \hline

Shome et al.~\cite{shome2021fast}
& Dual-arm
& Mixed-integer optimization
& Monotone (no intermediate buffering)
& Synchronous paired transfers
\\ \hline

Li et al.~\cite{li2024synchronized}
& Dual-arm
& Cooperative mTSP + RL
& Monotone (no temporal buffering)
& Shared-workspace synchronized actions
\\ \hline

Gao et al.~\cite{gao2024holistic}
& Dual-arm
& Integrated TAMP + trajectory optimization
& Lazy buffers
& Predefined handoff pose
\\ \hline

Barghi et al.~\cite{barghi2025dynamic}
& Single-arm
& A* + MCTS goal-attempting
& Movable stacking
& N/A
\\ \hline

\textbf{Proposed}
& \textbf{Dual-arm}
& \textbf{Joint MCTS}
& \textbf{Movable stacking}
& \textbf{Gap-Aware Push Transfer}
\\ \hline

\end{tabular}%
}
\end{table*}

Building on these observations, this paper presents a dual-arm tabletop rearrangement framework that combines Gap-Aware Push Transfer for cross-boundary exchange with movable stacking, as showed in Fig.~\ref{fig:overview}. Instead of restricting inter-arm transfer to a fixed aerial handoff pose, the proposed method searches the shared boundary for collision-free transfer gaps, places the object near a selected crossing point, and pushes it into the receiving workspace. This design reduces synchronous coupling between the two manipulators. In addition, a joint MCTS planner coordinates both arms while jointly reasoning about buffering and transfer decisions, and execution cost is used after planning to assess the quality of the generated plan. Compared with a fixed single-location handoff baseline, the proposed framework achieves better performance, particularly in denser scenes.

The proposed framework is detailed in Section~\ref{sec:framework}, including the problem formulation, Gap-Aware Push Transfer, the MCTS-based planner, and the experimental setup. The benchmark evaluation and discussion are presented in Section~\ref{sec:results}, and the conclusion is given in Section~\ref{sec:conclusion}.

\section{DUAL-ARM REARRANGEMENT FRAMEWORK}
\label{sec:framework}

This section presents the problem formulation and the proposed planning framework in a unified manner. The objective is to transform an initial tabletop arrangement into a desired target arrangement using two fixed-base manipulators while accounting for clutter, inter-arm reachability limits, movable stacking, and cross-boundary object transfer. Given the initial and target scenes, the framework generates a coordinated sequence of dual-arm actions that incrementally drives the scene toward the target configuration, as illustrated in Fig.~\ref{fig:pipeline_framework}.

\subsection{Problem Formulation}

Let $\mathcal{O}=\{o_1,\dots,o_n\}$ denote the set of objects to be rearranged. The tabletop workspace has dimensions $H \times W$. Each object $o_i$ is represented by an axis-aligned rectangular footprint of size $(h_i,w_i)$ with center position $p_i=(r_i,c_i)$, where $r_i$ and $c_i$ denote its row and column coordinates, respectively. Two fixed-base manipulators, $r_0$ and $r_1$, operate over adjacent halves of the workspace separated by a shared boundary column $c_b=\lfloor W/2 \rfloor$, which serves as the transfer interface between the two arms.

Following \cite{barghi2025dynamic}, the environment contains 12 heterogeneous object classes grouped according to their support properties: primary bases (e.g., box, pot, and basket), secondary bases (e.g., bowl, mug, and paper cup), lightweight items (e.g., spoon, fork, and knife), and heavier items (e.g., apple, pear, and banana). This categorization provides the basis for defining stacking relationships among objects, where objects with higher support capability can serve as bases for others. Consequently, the generated stacks remain physically plausible, and the stacking configurations are constrained to reduce the likelihood of instability during object manipulation or relocation.

To represent stacked arrangements, a directed stacking relation $R_t$ is maintained at time step $t$, where $R_t[i,j]=1$ indicates that object $o_i$ is directly stacked on top of object $o_j$. Let $P_t$ denote the object-placement component of the scene, consisting of the object centers together with their footprint attributes. The system state is then defined as
\[
s_t = (P_t, R_t, m_t^0, m_t^1),
\]
where $m_t^k$ denotes the current end-effector position of arm~$r_k$. A state is considered valid if all object footprints remain within workspace boundaries, no two distinct objects overlap on the tabletop plane except under admissible stacking relations, and all active support relations satisfy the predefined support-compatibility constraints. In addition, each manipulator may act only on objects located in its reachable side of the workspace, while the shared boundary interface remains accessible to both arms.

\begin{figure*}[t]
\centering
\includegraphics[width=\textwidth]{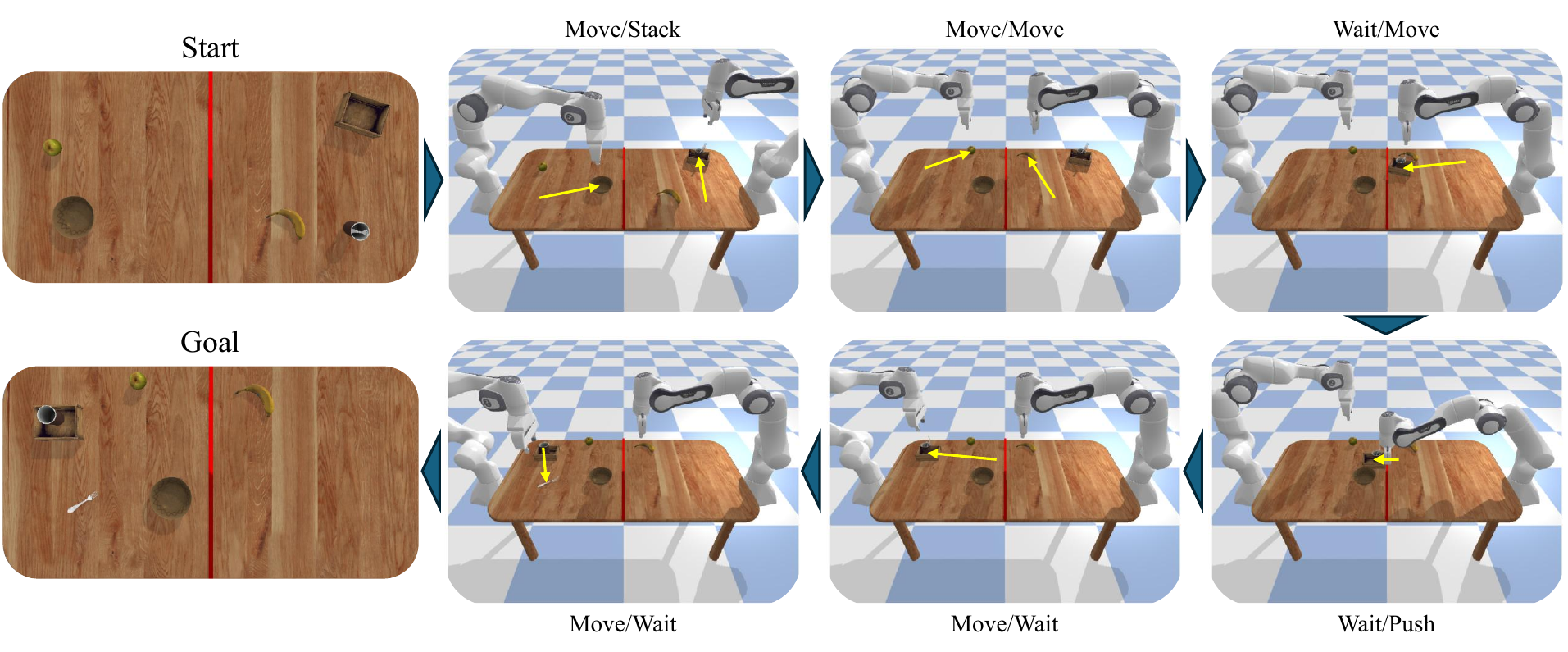}
\caption{Overview of the proposed dual-arm rearrangement framework, from initial and target scenes to coordinated dual-arm planning and execution. The figure also shows the primitive actions—\textit{move}, \textit{stack}, \textit{push}, and \textit{wait}—that form the basic operations for object manipulation and inter-arm coordination.}
\label{fig:pipeline_framework}
\end{figure*}

For the proposed planner, the primitive action set is defined as
\[
\mathcal{A}=\{\mathrm{move},\mathrm{stack},\mathrm{push},\mathrm{wait}\}.
\]
As shown in Fig.~\ref{fig:pipeline_framework}, the $\mathrm{move}$ action performs a prehensile pick-and-place relocation of an object to a collision-free placement and is used both for direct goal placement and for intermediate repositioning. The $\mathrm{stack}$ action places an object on a valid support object and is used whenever a temporary or target support relation must be established. The $\mathrm{push}$ action is reserved for cross-boundary transfer through the shared interface, and $\mathrm{wait}$ is used when no safe concurrent counterpart is available for the other manipulator.

At each time step, a joint action is written as
\[
a_t=(a_t^0,a_t^1),
\]
where $a_t^k \in \mathcal{A}$ is assigned to arm~$r_k$. Not every pair of actions is feasible. In particular, the two manipulators must not act simultaneously on the same source object, and concurrent stacking actions must not target the same support object~\cite{gao2022toward}. Moreover, even when the manipulated objects are distinct, motion interference or ordering dependencies may prevent safe parallel execution. In such cases, a valid single-arm action is paired with $\mathrm{wait}$ for the opposite arm.

In the present framework, temporary stacks may serve as movable stacking during candidate-action generation. When a base object is relocated, the objects stably supported on top of it are transported together while preserving their relative stacking relations~\cite{barghi2025dynamic}. This capability allows multiple objects to be moved temporarily as a single unit in cluttered scenes, thereby reducing unnecessary intermediate rearrangements.

\subsection{Coordinated Planning and Boundary Transfer}

The proposed planner uses a joint MCTS module as the core search mechanism~\cite{labbe2020monte}. In the present dual-arm setting, MCTS operates directly on feasible joint actions, enforces inter-arm compatibility constraints, incorporates movable stacking as an intermediate option during candidate generation~\cite{barghi2025dynamic}, and handles cross-boundary transfer decisions within the same search loop. In this sense, MCTS serves as a coordinated search mechanism for two-arm action selection under cluttered rearrangement constraints.

Each search-tree node corresponds to a state $s$, and each edge corresponds to a feasible joint action. During selection, child nodes are chosen according to an Upper Confidence Bound criterion. During expansion, candidate single-arm actions are generated for unresolved objects or for objects whose current placements obstruct direct progress toward the target arrangement. These actions are then combined into feasible joint actions subject to the coordination constraints defined above, with preference given to compatible parallel actions when available. During evaluation, a state is scored according to its progress toward the target arrangement, based on how many objects have already reached their desired placements or target support relations, together with a small preference for states that expose more immediate direct actions. The resulting value is then propagated upward during backup, completing the MCTS loop with the four stages of selection, expansion, evaluation, and backup.

\begin{figure}[t]
\centering
\includegraphics[width=\columnwidth]{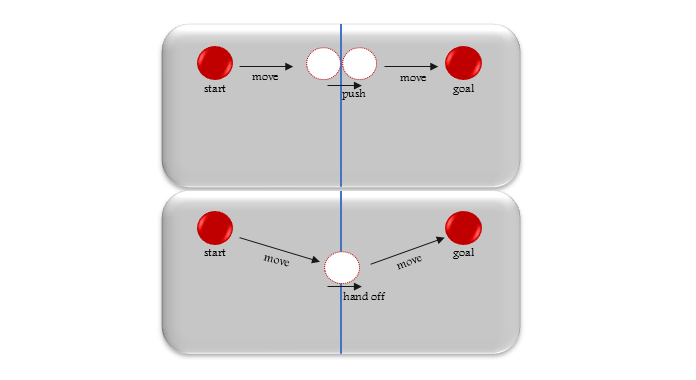}
\caption{Conceptual comparison between a fixed-handoff transfer scheme with a predefined transfer pose~\cite{gao2022toward} and the proposed method. The fixed point is illustrative, highlighting that the handoff location is predetermined rather than selected online.}
\label{fig:push_vs_handoff}

\end{figure}

Fixed-handoff formulations can be restrictive because cross-boundary exchange is tied to a predefined transfer pose, requiring the object to be routed through that configuration before it can enter the other workspace~\cite{gao2022toward,gao2024holistic}. As illustrated in Fig.~\ref{fig:push_vs_handoff}, the proposed framework instead uses a push-based transfer mechanism that can exploit feasible boundary intervals compatible with the current scene geometry. This provides greater flexibility than a single fixed handoff pose and can shorten the transfer path when a nearby crossing location is available.

\begin{figure}[t]
\centering
\includegraphics[width=\columnwidth]{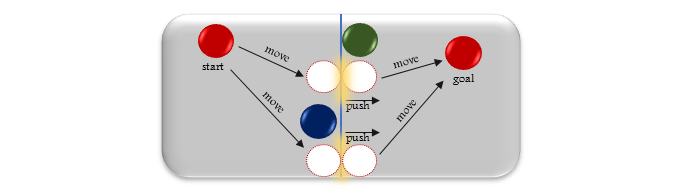}
\caption{Selection of the crossing row from feasible free intervals along the boundary strip.}
\label{fig:gap_selection}

\end{figure}

Once a push-based transfer is selected, the planner examines the boundary strip on both sides of the shared interface to identify feasible crossing intervals for the object, as depicted in Fig.~\ref{fig:gap_selection}. In practice, only intervals that provide sufficient clearance for both the object footprint and the pre-push alignment maneuver are considered admissible. Among the resulting candidate crossing locations, the planner prefers those that remain close to the current object row and, when relevant, also to the target row.

Let $\mathcal{G}(o_i)$ denote the set of valid candidate crossing rows for object $o_i$. The selected crossing row is then defined as
\[
r_i^{\mathrm{cross}}=
\argmin_{\rho \in \mathcal{G}(o_i)}
\left(
|\rho-r_i|+|\rho-r_i^*|
\right)
\]
where $r_i$ is the current row of object $o_i$, $r_i^*$ is its target row when relevant, and $\rho$ denotes a candidate crossing row along the shared boundary. If the target row is not involved in the transfer decision, the criterion reduces to selecting the valid candidate closest to the current row. In this way, the chosen crossing location remains compatible with the local scene geometry while avoiding unnecessary deviation from both the current placement and the intended final arrangement, as shown in Fig.~\ref{fig:gap_selection}.

Accordingly, a push transfer is executed in three stages: first, a prehensile alignment move places the object near the selected crossing row; second, a straight push carries the object across the shared boundary; and finally, the object becomes available for subsequent manipulation by the receiving arm~\cite{dogar2011framework}. By describing the transfer in terms of functional stages rather than as a rigid action sequence, flexibility is retained in the scheduling of the corresponding motions within the joint plan. Consequently, the crossing location is treated as a planning variable rather than being prescribed a priori as a fixed design choice.

The MCTS planner is progress-driven rather than explicitly cost-minimizing. During search, states are evaluated according to their progress toward the target arrangement, while execution cost is used after planning to assess the quality of the resulting plan. For an action sequence
\[
\Pi=(a_0,\dots,a_{T-1}),
\]
the execution cost is defined as
\[
J(\Pi)=\sum_{t=0}^{T-1} C(a_t),
\]
where $T$ is the plan length, $a_t$ is the action executed at step $t$, and $C(a_t)$ denotes the execution cost of that action.

For $\mathrm{move}$ and $\mathrm{stack}$, the action cost includes manipulator travel to the manipulated object, transport to the resulting placement, and a fixed primitive-execution term. For a push-based boundary transfer, the cost accounts for the full transfer maneuver, including approach to the object, alignment near the selected crossing row, the straight push across the shared boundary, and the retreat motion after the push. Thus, the cost reflects the overall motion and manipulation effort required by the executed action rather than only the boundary-crossing displacement itself.

Accordingly, $J(\Pi)$ serves as a post-planning measure of execution effort and is used to evaluate the generated plans and compare the considered transfer strategies in the benchmark.

\section{RESULTS AND DISCUSSION}
\label{sec:results}

In this section, the proposed framework is evaluated through a comparative benchmark against an MCTS-based implementation of the fixed-handoff formulation introduced by~\cite{gao2022toward}. By using the same search backbone for both methods, the comparison considered the effect of the transfer mechanism itself, namely, fixed handoff versus dynamic Gap-Aware Push Transfer. The two methods are compared in terms of Joint Steps and Execution Cost under increasing scene density.

All experiments are conducted on a tabletop workspace of size $1 \times 2$ $m$, with a shared boundary located at the middle column. The number of objects is denoted by $N$, and the benchmark considers $N \in \{5,6,7,8,9,10,11,12\}$ to evaluate performance under different clutter levels. For each value of $N$, $20$ random scene pairs are generated, each consisting of an initial configuration and a corresponding goal configuration. Each method is executed $3$ times for every scene pair to reduce stochastic variation in the search process.

The benchmark is conducted using heterogeneous object sizes during scene generation, following the general setting adopted in prior stacking-based rearrangement experiments~\cite{barghi2025dynamic}. Rather than using a uniform object set, each scene contains objects of different sizes, yielding more diverse rearrangement instances and a broader evaluation of the planning framework under non-uniform clutter.

Two methods are considered: The Fixed Handoff baseline applies the fixed handoff concept by limiting cross-boundary exchanges to a single predefined transfer pose~\cite{gao2022toward}. Although an MCTS-based dual-arm planner was implemented for consistency, the benchmark contrasts this fixed approach with the proposed Gap-Aware Push Transfer method, which dynamically selects transfer opportunities based on the scene configuration.

\begin{figure}[t]
    \centering
    \includegraphics[width=\linewidth]{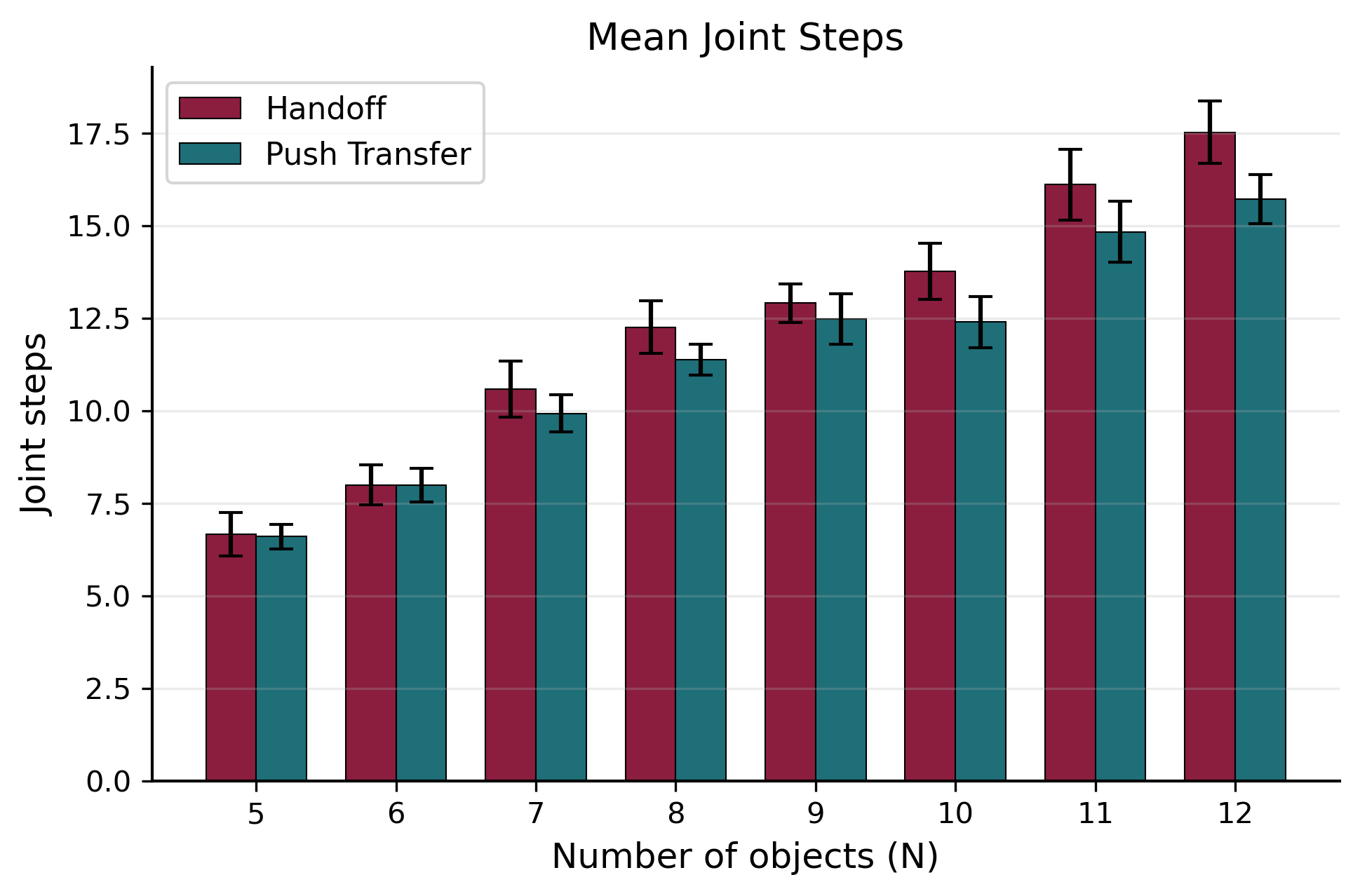}

    \vspace{0.4em}

    \includegraphics[width=\linewidth]{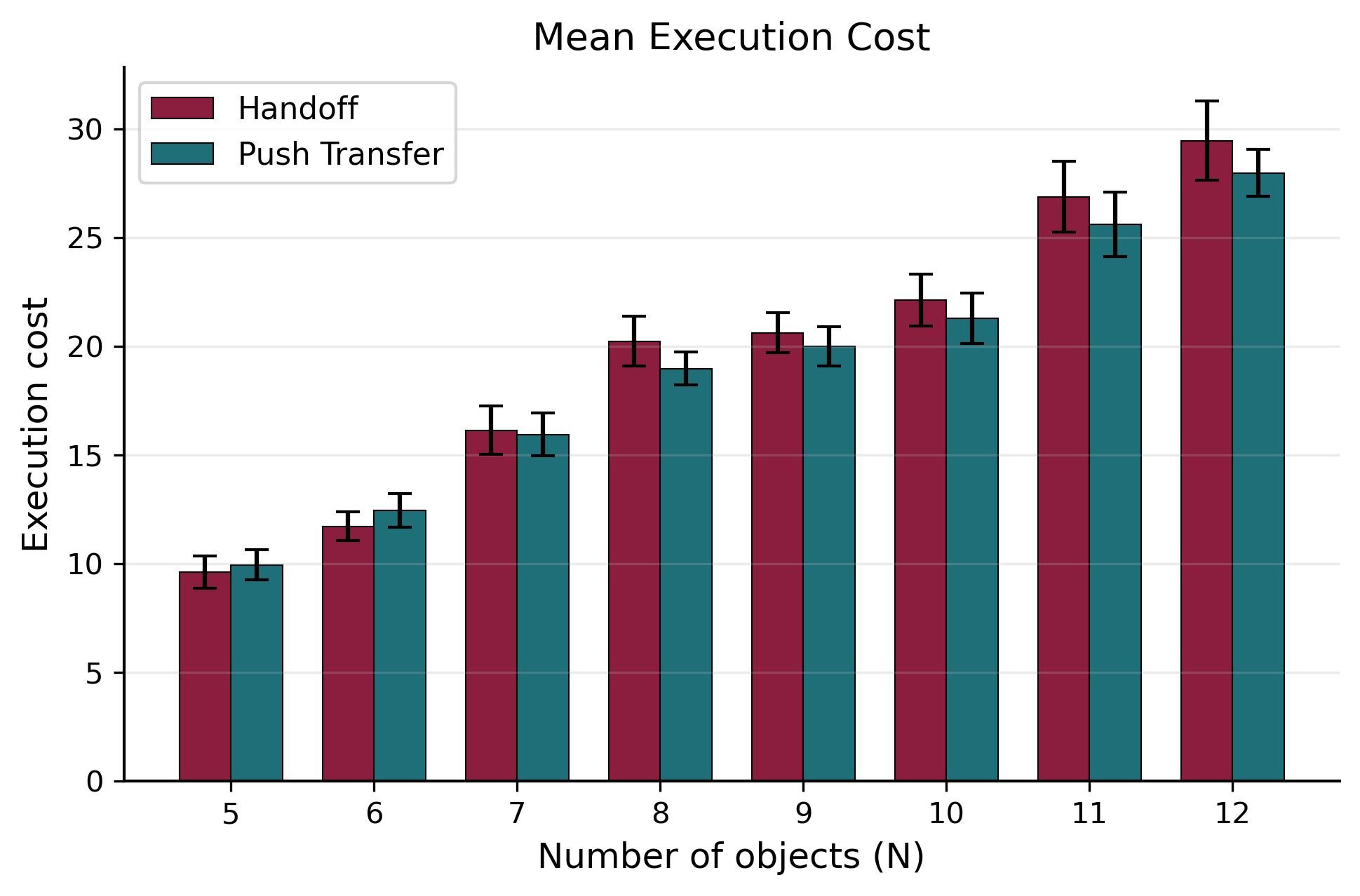}

    \caption{Performance comparison of two MCTS-based strategies across varying object counts. Bars represent mean values with confidence intervals. The proposed method demonstrates a clearer advantage over the Fixed Handoff baseline.}
    \label{fig:metrics_combined}
\end{figure}

The benchmark is evaluated using two metrics: Joint Steps and Execution Cost. Joint Steps measure the number of coordinated execution steps required to complete the rearrangement. Execution Cost is computed as the cumulative post-planning action cost of the generated plan, as defined in Section~II. Together, these metrics characterize coordination efficiency and overall execution effort under increasing scene density. The aggregated quantitative results are reported in Table~\ref{tab:main_results}, and Fig.~\ref{fig:metrics_combined} illustrates how the relative behavior of the two strategies changes as the number of objects increases.

\begin{table}[t]
\caption{\centering Benchmark results for the compared methods under increasing scene density. The advantage of the proposed strategy becomes more pronounced in denser scenes.}
\label{tab:main_results}
\centering
\resizebox{\columnwidth}{!}{%
\begin{tabular}{c l c c}
\hline
$N$ & Strategy & Joint Steps & Execution Cost \\
\hline
\multirow{2}{*}{5}  & Fixed Handoff~\cite{gao2022toward}           & 6.67 & 9.63 \\
                    & Proposed (Gap-Aware Push Transfer)          & 6.60 & 9.96 \\
\hline
\multirow{2}{*}{6}  & Fixed Handoff~\cite{gao2022toward}           & 8.00 & 11.73 \\
                    & Proposed (Gap-Aware Push Transfer)          & 7.98 & 12.47 \\
\hline
\multirow{2}{*}{7}  & Fixed Handoff~\cite{gao2022toward}           & 10.58 & 16.15 \\
                    & Proposed (Gap-Aware Push Transfer)          & 9.93 & 15.96 \\
\hline
\multirow{2}{*}{8}  & Fixed Handoff~\cite{gao2022toward}           & 12.26 & 20.25 \\
                    & Proposed (Gap-Aware Push Transfer)          & 11.38 & 18.99 \\
\hline
\multirow{2}{*}{9}  & Fixed Handoff~\cite{gao2022toward}           & 12.90 & 20.63 \\
                    & Proposed (Gap-Aware Push Transfer)          & 12.48 & 20.02 \\
\hline
\multirow{2}{*}{10} & Fixed Handoff~\cite{gao2022toward}           & 13.77 & 22.14 \\
                    & Proposed (Gap-Aware Push Transfer)          & 12.40 & 21.30 \\
\hline
\multirow{2}{*}{11} & Fixed Handoff~\cite{gao2022toward}           & 16.11 & 26.90 \\
                    & Proposed (Gap-Aware Push Transfer)          & 14.83 & 25.64 \\
\hline
\multirow{2}{*}{12} & Fixed Handoff~\cite{gao2022toward}           & 17.52 & 29.48 \\
                    & Proposed (Gap-Aware Push Transfer)          & 15.72 & 27.99 \\
\hline
\end{tabular}%
}

\end{table}

As reported in Table~\ref{tab:main_results} and Fig.~\ref{fig:metrics_combined}, the relative behavior of the two methods varies with scene density. For smaller scenes, the two methods remain close. At $N=5$, the proposed method reduces Joint Steps from $6.67$ to $6.60$ (about $1.0\%$), while its Execution Cost increases from $9.63$ to $9.96$ (about $3.4\%$). A similar near-parity is observed at $N=6$, where the difference in Joint Steps is only $0.02$, and Fixed Handoff still retains a lower Execution Cost.

As the number of objects increases, however, the trend shifts in favor of the proposed method. From $N=7$ onward, Gap-Aware Push Transfer yields lower values for both Joint Steps and Execution Cost. At $N=7$, Joint Steps decrease from $10.58$ to $9.93$ (about $6.1\%$), while Execution Cost decreases from $16.15$ to $15.96$ (about $1.2\%$). The same tendency is visible across the intermediate dense cases. The separation becomes clearer in denser scenes. At $N=12$, Joint Steps decrease from $17.52$ to $15.72$ (about $10.3\%$), and Execution Cost decreases from $29.48$ to $27.99$ (about $5.1\%$). These quantitative differences indicate that the benefit of scene-dependent cross-boundary transfer becomes more pronounced as the shared boundary becomes increasingly contested.

Since both methods share the same MCTS backbone and stacking capability, so performance differences stem from the transfer mechanism. The proposed method selects scene-dependent crossing locations, while Fixed Handoff restricts exchange to a predefined pose. This flexibility reduces detours in dense scenes, lowering Joint Steps and Execution Cost.

\enlargethispage{5\baselineskip}
\section{Conclusion}
\label{sec:conclusion}
This paper presented a dual-arm tabletop rearrangement framework that combines movable stacking with Gap-Aware Push Transfer for cross-boundary exchange. Integrated within a joint MCTS planner, these components enable transfer decisions to be made dynamically rather than through a fixed handoff configuration. Experimental results showed that although Fixed Handoff remains competitive in lighter scenes, the proposed method achieves lower Joint Steps and lower Execution Cost as scene density increases. These results highlight the benefit of scene-dependent transfer selection for dense dual-arm tabletop rearrangement.

\end{document}